\documentclass[conference]{IEEEtran}
\IEEEoverridecommandlockouts
\usepackage{cite}
\usepackage{amsmath,amssymb,amsfonts}
\usepackage{algorithmic}
\usepackage{graphicx}
\usepackage{textcomp}
\usepackage{xcolor}
\usepackage{framed,multirow}
\usepackage{amsthm}

\usepackage{amssymb}
\usepackage{latexsym}
\usepackage{changepage} % 用于调整页边距
\usepackage{lipsum} % 用于生成示例文本
\usepackage{url}

\usepackage[hidelinks,colorlinks,urlcolor=gray,citecolor=gray,linkcolor=gray]{hyperref}
\usepackage{graphicx}
\usepackage{amsmath}
\usepackage{multicol}
\usepackage{amsfonts}
\usepackage{booktabs}
\usepackage{siunitx}
\usepackage{marvosym}

\def\BibTeX{{\rm B\kern-.05em{\sc i\kern-.025em b}\kern-.08em
    T\kern-.1667em\lower.7ex\hbox{E}\kern-.125emX}}
\begin{document}

\title{Achieve Fairness without Demographics for Dermatological Disease Diagnosis
}

\author{\IEEEauthorblockN{Ching-Hao Chiu}
\IEEEauthorblockA{\textit{Department of Computer Science} \\
\textit{National Tsing Hua University}\\
Hsinchu, Taiwan \\
gwjh101708@gapp.nthu.edu.tw}
\and
\IEEEauthorblockN{Yu-Jen Chen}
\IEEEauthorblockA{\textit{Department of Computer Science} \\
\textit{National Tsing Hua University}\\
Hsinchu, Taiwan \\
yujenchen@gapp.nthu.edu.tw}
\and

\IEEEauthorblockN{Yawen Wu}
\IEEEauthorblockA{\textit{Department of Computer Science and Engineering} \\
\textit{University of Notre Dame}\\
IN, USA \\
yawen.wu@pitt.edu}
\and
% \linebreakand %
\IEEEauthorblockN{Yiyu Shi}
\IEEEauthorblockA{\textit{Department of Computer Science and Engineering} \\
\textit{University of Notre Dame}\\
IN, USA \\
yshi4@nd.edu}
\and
\IEEEauthorblockN{Tsung-Yi Ho}
\IEEEauthorblockA{\textit{Department of Computer Science and Engineering} \\
\textit{The Chinese University of Hong Kong}\\
Hong Kong, People’s Republic of China \\
tyho@cse.cuhk.edu.hk}
}

\maketitle

\begin{abstract}
In medical image diagnosis, fairness has become increasingly crucial. Without bias mitigation, deploying unfair AI would harm the interests of the underprivileged population and potentially tear society apart. Recent research addresses prediction biases in deep learning models concerning demographic groups (e.g., gender, age, and race) by utilizing demographic (sensitive attribute) information during training. However, many sensitive attributes naturally exist in dermatological disease images. If the trained model only targets fairness for a specific attribute, it remains unfair for other attributes. Moreover, training a model that can accommodate multiple sensitive attributes is impractical due to privacy concerns. To overcome this, we propose a method enabling fair predictions for sensitive attributes during the testing phase without using such information during training. Inspired by prior work highlighting the impact of feature entanglement on fairness, we enhance the model features by capturing the features related to the sensitive and target attributes and regularizing the feature entanglement between corresponding classes. This ensures that the model can only classify based on the features related to the target attribute without relying on features associated with sensitive attributes, thereby improving fairness and accuracy. Additionally, we use disease masks from the Segment Anything Model (SAM) to enhance the quality of the learned feature. Experimental results demonstrate that the proposed method can improve fairness in classification compared to state-of-the-art methods in two dermatological disease datasets.
\end{abstract}

\begin{IEEEkeywords}
Dermatological Disease Diagnosis, AI Fairness, Fairness through Unawareness
\end{IEEEkeywords}

\section{Introduction}
\label{sec:intro}
In recent years, many institutions have introduced machine learning-based medical diagnostic systems. While these systems have achieved high accuracy in predicting disease conditions, there are biases in predicting results for different population groups in skin disease datasets, as shown in \cite{groh2021evaluating, kinyanjui2020fairness, tschandl2018ham10000}. Predictive biases can occur when there is an imbalance in the number of disease images of different demographic groups, leading to inaccurate predictions and misdiagnoses. The discriminatory nature of these models can harm society, causing distrust in computer-assisted diagnostic methods among other sensitive groups, such as race or gender.

Recently, various methods have been proposed to mitigate bias in machine learning models. Many methods \cite{wang2022fairness, zhang2018mitigating, kim2019learning, ngxande2020bias} use adversarial training to train networks to learn classifiers while removing the adversary's ability to classify sensitive attributes to eliminate bias. Another mainstream approach is regularization-based methods, such as \cite{quadrianto2019discovering, jung2021fair}, which use specific loss functions to constrain the model to learn the features unrelated to sensitive attributes. Contrastive learning \cite{park2022fair} and pruning \cite{wu2022fairprune, lin2022fairgrape} have also received extensive research in recent years. These methods mitigate predictive biases in classification by using demographic information (sensitive attributes, e.g., gender, race, age) in the training data.

However, collecting sensitive attribute information is only sometimes feasible due to privacy or legal issues. The methods that can achieve fair prediction without sensitive attributes are proposed to overcome this issue. Mainstream methods inspired by the max-min problem include distributionally robust optimization (DRO) \cite{hashimoto2018fairness}, adversarial learning-based methods \cite{lahoti2020fairness}, and fair self-supervised learning \cite{chai2022self}. However, these methods often significantly reduce the model's accuracy. There are other methods, such as knowledge distillation \cite{chai2022fairness} or multi-exit training \cite{chiu2023toward, chiu2023fair}. However, further research is needed for the knowledge distillation method due to a limited understanding of the generality and quality of the knowledge learned, particularly in measuring knowledge quality related to fairness concepts \cite{gou2021knowledge}. As for multi-exit training, the impact on predictive performance may vary due to changes in network architecture, and improvement is only sometimes guaranteed \cite {chiu2023fair}. Extending this framework to other methods is needed to enhance fair classification outcomes.

\begin{figure}[!h]
\setlength{\belowcaptionskip}{-5pt} 
% \vspace*{-4cm}
% \begin{center}

% width=0.9\linewidth
\includegraphics[width=1.0\linewidth]{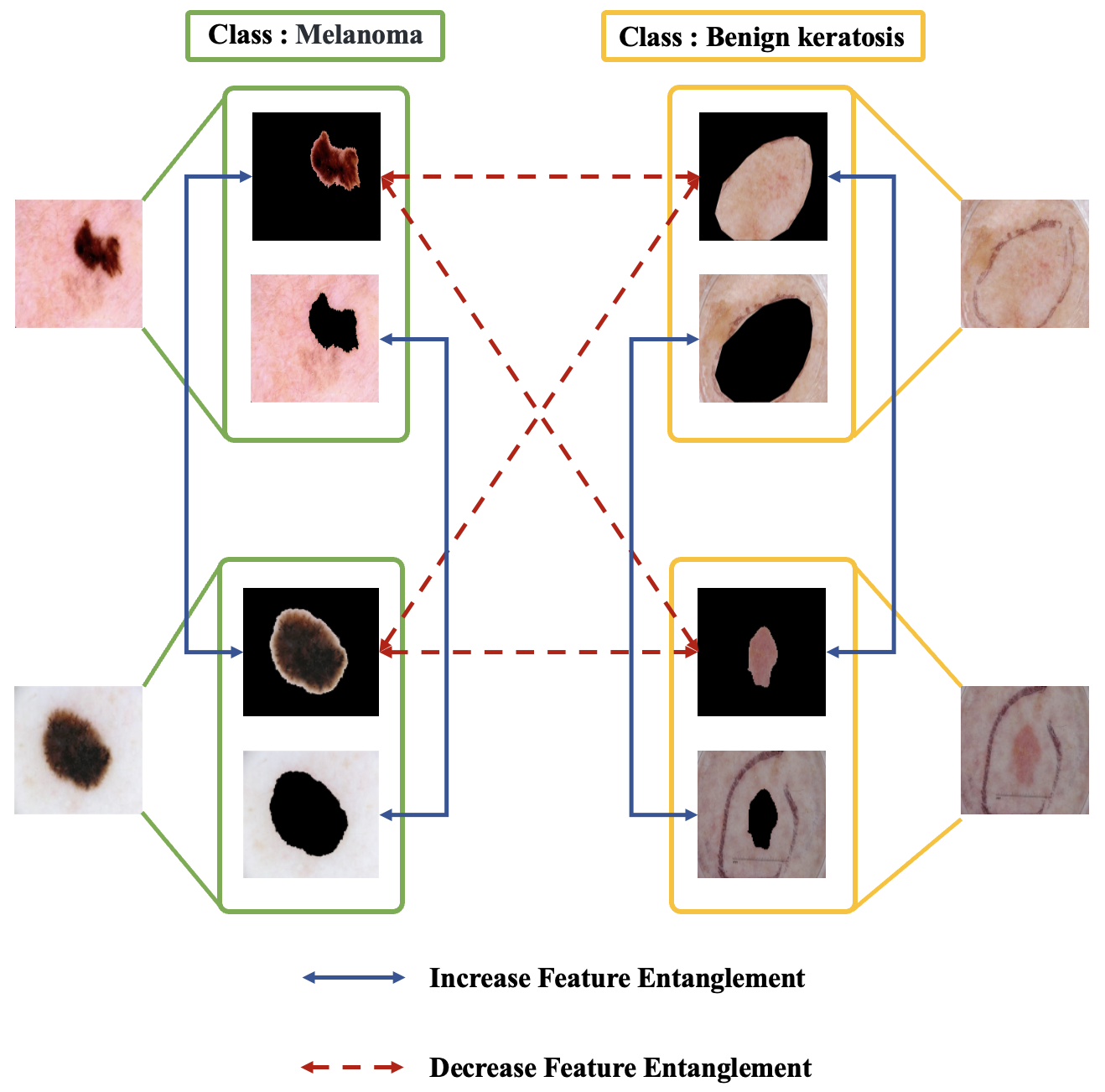}
\caption{The concept of our motivation for the training framework. The image that masks the skin part represents the feature map related to the diseased part, and the image that masks the diseased part signifies the feature map related to the skin part.}
\label{Fig.Concept}
% \end{center}
\end{figure}

In this paper, we are inspired by the observation in \cite{chiu2023toward}, which found that models can achieve fairer predictions by using features highly entangled with different sensitive attribute classes for classification. Using the soft nearest neighbor loss (SNNL) as a measure, we can assess the degree of entanglement between different class features in the embedding space, as introduced in \cite{frosst2019analyzing}. If the SNNL values across different sensitive attribute classes are low, features become more distinguishable among sensitive attributes, and if they are high, they become indistinguishable. Since we aim to improve predictive fairness across different demographic groups without using sensitive attribute information, we designed a training framework. This framework ensures that the model classifies disease types only based on features related to the diseased part (target attribute) without distinguishing features related to the skin (sensitive attribute) within the same disease category. This design aims to prevent the model from learning to classify diseases based on skin features, which can lead to predictive performance biases among different demographic groups in the same disease category. Specifically, by using attention modules to obtain features that focus on the diseased part and the skin part and then using SNNL as a regularization constraint, the model's ability to differentiate diseases is enhanced, while its ability to determine skin differences within the same disease category is reduced. Our concept is shown in Fig. \ref{Fig.Concept}, where the red dashed arrow represents decreased feature entanglement between different features, and the blue solid line represents increased feature entanglement. For images belonging to the same class, such as those on the left belonging to the ``Melanoma'' class, the feature entanglement related to the diseased part and skin part should be increased between different images; for images from other classes, such as ``Benign keratosis'' shown in the figure, the feature entanglement relates to the diseased part should be decreased. To improve the quality of attention maps during training, we also incorporate disease masks generated by the Segment Anything Model (SAM) \cite{kirillov2023segment} into the obtained features, further enhancing the model's accuracy and fairness.

Our extensive experiments demonstrate that our proposed \textbf{Att}ention-based feature \textbf{EN}tanglement regularization method (\textbf{AttEN}) can achieve fairness without using sensitive attributes. This method is more suitable for skin disease diagnosis than previous bias mitigation methods because sensitive attribute information may involve privacy issues and may not always be available. This method offers better predictive accuracy and fairness than other methods that do not require collecting sensitive attribute information and can maintain robust results across different model architectures. We compare our method with the current state-of-the-art methods that require using sensitive attributes during training and methods that do not require sensitive attributes, showing the best overall performance in these comparisons.

The main contributions of the proposed method are as follows:

\begin{itemize}
    \item We propose a method to improve predictive fairness in dermatological disease classification without using sensitive attribute information.
    \item With theoretical justification, we have confirmed that our approach of capturing features and performing entanglement regularization through SAM and the attention module can enhance the fairness of the learned features.
    \item Through extensive experiments, we show that our approach can improve fairness while maintaining competitive accuracy on both the dermatological disease datasets, ISIC 2019 \cite{combalia2019bcn20000}, and Fitzpatrick-17k datasets \cite{groh2021evaluating}.

\end{itemize}

\section{Background and Related Work}
\label{sec:background_related}
\subsection{Fairness Criterion}
Consider the predictor $F$, the input variable $X$, the output ground truth variable $Y$, and the predictor output variable $\hat{Y} = F(X)$. Regarding the sensitive attribute variable $A$, there are two definitions of fairness.

\textbf{Definition 1.} \textit{(Equalized Opportunity \cite{hardt2016equality})} A predictor $F$ satisfies equalized opportunity for a class $y$ if $\hat{Y}$ and $A$ are independent conditioned on $Y = y$. That is, for a particular true label value $y$ , $P(\hat{Y} = \hat{y})$ is the same for all values of the variable $A$:

\begin{equation}
\label{eq.EOpp_def_1}
P(\hat{Y}=\hat{y} \mid Y=y) = P(\hat{Y}=\hat{y} \mid A=a,Y=y).
\end{equation}

\textbf{Definition 2.} \textit{(Equalized Odds \cite{hardt2016equality})} A predictor $F$ satisfies equalized odds if $\hat{Y}$ and $A$ are conditionally independent given $Y$. That is, for all possible values of the true label value $y$ , $P(\hat{Y} = \hat{y})$ is the same for all values of the variable $A$:
\begin{equation}
\label{eq.EO_def_1}
P(\hat{Y}=\hat{y} \mid Y=y) = P(\hat{Y}=\hat{y} \mid A=a,Y=y).
\end{equation}

To evaluate the fairness performance of the training framework, the fairness metric, equalized opportunity (Eopp0 and Eopp1), and equalized odds (Eodd) can be formalized based on the Eq.(\ref{eq.EOpp_def_1}) and Eq.(\ref{eq.EO_def_1}) \cite{hardt2016equality}. The Eopp0, Eopp1, and the Eodd are computed as below equations, where $TPR_{k}^{a}$, $TNR_{k}^{a}$ and $FPR_{k}^{a}$ are the True Positive Rate, True Negative Rate and the False Positive Rate respectively of target class $k$ and sensitive attribute $a$, in this paper, we only consider the binary sensitive attribute (i.e., $a \in \{0, 1\}$ ) : 

\begin{equation}
\label{eq.EOpp0_equation}
    Eopp0 = \sum_{k=1}^{K} |TNR_{k}^{1} - TNR_{k}^{0}|
\end{equation}

\begin{equation}
\label{eq.EOpp1_equation}
    Eopp1 = \sum_{k=1}^{K} |TPR_{k}^{1} - TPR_{k}^{0}|
\end{equation}

\begin{equation}
\label{eq.EO_equation}
    Eodd = \sum_{k=1}^{K} |TPR_{k}^{1} - TPR_{k}^{0} + FPR_{k}^{1} - FPR_{k}^{0}|
\end{equation}
The equalized opportunity considers the true negative rate and true positive rate difference between different sensitive groups, while the equalized odds consider the true positive rate and false positive rates between different sensitive groups. The model keeps lower equalized opportunity and equalized odds mean the model is fairer.

\subsection{Related Work}

\subsubsection{Bias Mitigation Method}
Bias mitigation methods are designed to reduce the native bias in the dataset to reduce the chance of unfair prediction. In the recent study, the debias strategy can be divided into two groups: fairness with and without demographics.

\textbf{Fairness with demographic} methods involves using sensitive attributes information in their debiasing strategy, for example, removing information that may cause discrimination from training data before training \cite{ngxande2020bias,lu2020gender}, or assigning different weights to different data samples to suppress sensitive information during training \cite{kamiran2012data}. Another approach is to modify off-the-shelf model architecture, training strategy, and loss functions to achieve fairness goals. This approach includes adversarial training \cite{zhang2018mitigating, kim2019learning, wang2022fairness}, which mitigates bias by adversarially training an encoder and classifier to learn a fair representation. Some regularization-based methods, such as those mentioned in \cite{quadrianto2019discovering}, use the Hilbert-Schmidt norm to know a fair representation that retains the features' semantics from the input domain. \cite{jung2021fair} learned a fair representation by distilling the fair information of the teacher model into the student model using the Maximum Mean Discrepancy (MMD) loss \cite{gretton2012kernel}. \cite{park2022fair} introduced group-wise normalization and penalized the inclusion of sensitive attributes to mitigate the biased condition of contrastive learning. Furthermore, there exist methods aime at enhancing fairness by adjusting the model's output distribution to match the training distribution or a specific fairness metric. This calibration is done by taking the model's predictions and the sensitive attribute as inputs \cite{hardt2016equality, zhao2017men, du2020fairness}. Beyond these methodologies, fairness has also been sought in the domain of dermatological disease diagnosis, with techniques like pruning \cite{wu2022fairprune} and batch normalization \cite{xu2023fairadabn} coming to the forefront. 

However, these techniques require sensitive attributes in their training process, which is only sometimes suitable for medical diagnosis tasks due to privacy issues. Additionally, since these methodologies typically target fairness constraints specific to certain sensitive attributes, deploying a model adaptable to all sensitive attributes may necessitate retraining to align with new objectives, potentially compromising fairness in other attributes.

\textbf{Fairness without demographic} approaches are proposed due to limitations in collecting demographic information. When demographic information is inaccessible, methods to achieve fairness without prior knowledge of the sensitive attributes have emerged. Some approaches are inspired by the Rawlsian Maximin principle \cite{rawls2001justice}, such as the distributionally robust optimization (DRO)-based approach \cite{hashimoto2018fairness} and the adversarial learning-based approach \cite{lahoti2020fairness}. Other methods, such as using input features to find surrogate subgroup information \cite{zhao2022towards}, knowledge distillation \cite{chai2022fairness}, have also been proposed. However, these approaches may generally result in a high-performance degradation of the model's accuracy.

Recently, in \cite{chiu2023toward}, the authors leveraged the fair-multi exit framework without sensitive attribute information during training in dermatological disease diagnosis to achieve fair prediction. They aim to classify highly entangled features with sensitive groups to improve prediction fairness. During inference time, for shallow features, if the confidence in classification is high enough, the model can exit early, using the classification result as the final prediction. However, relying solely on confidence scores cannot explain whether the features learned by the model are fair. In other words, there needs to be a specific metric to quantify the fairness of shallow features. Moreover, the performance of using shallow features for classification is not optimal. As a result, this framework in a vanilla CNN cannot significantly improve; thus, applying their framework to other state-of-the-art models, which need sensitive attribute information, is still necessary to achieve a better fairness score in this task.

In this paper, our focus is on improving fairness in dermatological disease diagnosis without sensitive attribute information. By designing a regularization loss that aligns with the classification objective and introducing a high-confidence mask generated by SAM, we can further enhance predictive fairness and accuracy while preserving patient privacy.

\subsubsection{Segment Anything Model}
The Segment Anything Model (SAM) is built on the largest segmentation dataset, SA-1B, which comprises over 11 million images with over 1 billion ground-truth segmentation masks. SAM excels in zero-shot applications, meaning it achieves high accuracy without retraining or fine-tuning on new, unseen datasets or segmentation tasks, outperforming other interactive or dataset-specific models.

SAM's architecture consists of three main components: image encoder, prompt encoder, and mask decoder. The image encoder uses the same architecture as the Vision Transformer (ViT) \cite{dosovitskiy2020image} pre-trained by MAE \cite{he2022masked}. Three different scales image encoders, ViT-H, ViT-L, and Vit-B, allow a trade-off between real-time performance and accuracy.

The image encoder can accept input images of any size and reshape them into a 1024x1024 format. The images are then converted into serialized patch embeddings with a patch size of 16x16 and an embedding size 256. After passing through multiple Transformer blocks with window attention and residual propagation, the output dimensions of the image encoder are (64x64, 256).

The prompt encoder supports four prompts: point, box, mask, and text. Point and box prompts can be used as a single category with position encodings, while text prompts can use CLIP as an encoder. These are all sparse prompts projected onto prompt tokens and concatenated with image embeddings. In contrast, dense prompts, like masks, are embedded using convolutions and summed element-wise with the image embeddings.

The mask decoder uses prompt self-attention and cross-attention in two directions (prompt-to-image embedding and vice-versa) to update all embeddings. After running two blocks, upsample the image embedding, and an MLP maps the output token to a dynamic linear classifier, which then computes the mask foreground probability at each image location.

Several studies have been on fine-tuning SAM for medical image applications, including \cite{wu2023medical, he2023accuracy, roy2023sam}. To the best of our knowledge, our work is the first to apply SAM to improve fairness in medical diagnosis, addressing fairness without demographics in dermatological disease diagnosis.

% \begin{figure*}[!h]
% \setlength{\belowcaptionskip}{-15pt} 
% \begin{center}

% \includegraphics[width=1.0\linewidth]{Fig/Atten_Method.png}
% \caption{Illustration of the proposed ``AttEN'' training framework. The upper arrow entering the AttEN module represents the original input images, while the lower arrow entering AttEN represents the features from previous internal layers, denoted as \textit{Feature $A$}. The arrow from the AttEN module, marked as \textit{Feature $A^{'}$}, will proceed to the next layer in the model. This module can be appended at any layer in the neural network; it can also append multiple modules in the neural network.
% }
% \label{Fig.Method}
% \end{center}
% \end{figure*}

\section{Method}
\label{sec:method}
\subsection{Problem Formulation}
\label{sec:method_problem}

In the classification task, we define the input features as $x \in X = \mathbb{R}^{d}$, the target class as $y \in Y = \{1, 2, \ldots, N\}$, and the sensitive attributes as $a \in A = \{1, 2, \ldots, M\}$. The goal is to train a classifier $F : X \rightarrow Y$ that predicts the target class $y$ to achieve high accuracy while remaining unbiased for the sensitive attributes $a$. In this paper, we specifically focus on the scenario where the sensitive attributes are unknown during the training process and are only used to assess fairness performance during testing time.

\subsection{Avoid Shortcuts Learning in Bias Dataset}
The motivation behind designing our method is to achieve ``Group Fairness'' \cite{dwork2012fairness, verma2018fairness}, which means the model has consistent predictive performance across different sensitive groups. To achieve this, excluding features unrelated to the diseased part as much as possible during classification is essential, preventing the model from learning classification through shortcuts \cite{geirhos2020shortcut}. This is because the shortcuts in the model prediction will cause bias prediction. Here, we will justify how our method improves fairness by avoiding model learning shortcuts.    

% given class k，number belong to sensitive a is sensitive a'
% the model will tend to learn ...

% For a dataset with $N$ classes, if there is a class $k$ with a significantly higher count in sensitive attribute $a$ compared to sensitive attribute $a'$, leading the model to learn how to classify the class $k$ based on whether or not the image has the sensitive attribute $a$. It leads to shortcut learning; that is, the model learns that an image belongs to $k$ if it has the sensitive attribute $a$ and vice versa. This situation leads to two types of biases in the model's predictions:    

In class $k$, if the number of images belonging to sensitive group $A$ with sensitive attribute $a$ is significantly more than those belonging to sensitive group $A'$ with sensitive attribute $a'$, the model has a higher probability of classifying an image belonging to class $k$ if it has the sensitive attribute $a$. Conversely, it reduces the likelihood of predicting an image belonging to class $k$ if it learns that the image does not have the sensitive attribute $a$. This scenario introduces two types of biases in the model's predictions:

% 如果學到a'，則較不可能是k

% 邏輯有點問題
% 非k來說 a' > a
% it's lead to shortcut learning 可以放上面那段

% a作為一種補充資訊，使得屬於其他class的image，如果被判斷出屬於a，則屬於class k?

% 屬於其他class的image，如果也學到了a對應該class的資訊呢？

% 邏輯錯誤在於如果學到a，不見得他就是屬於k

% 如果我們再多假設，在其他class中，並沒有學到sensitive attribute 為 a，則屬於該class的shortcut

% 如果改成學到 a ，則判斷成 k的機率提高 -> 如果學到a'，則判斷成k的機率降低

\begin{enumerate}
    \item For Group $A$ with sensitive attribute $a$, element not belongs to class $k$ are more likely to be predicted as class $k$, and increase the false positive ($FP$) prediction cases.
    \item For Group $A'$ with sensitive attribute $a'$, element belongs to class $k$ are more likely to be predicted as not belongs to $k$, and increase the false negative ($FN$) prediction cases.
\end{enumerate}

% During the inference stage, for a dataset with consistent counts for both sensitive groups in all classes, an unbiased model predicting all cases for $k$ would have \(TP_A = TP_{A'}, FP_A = FP_{A'}, TN_A = TN_{A'}, FN_A = FN_{A'}\) (Here we consider the counts for both sensitive groups in consistent for clear representation, if the count for different sensitive group are different, these term can easily substitute to \(TP_A = \alpha*TP_{A'}, FP_A = \alpha*FP_{A'}, TN_A = \alpha*TN_{A'}, FN_A = \alpha*FN_{A'}\), where $\alpha$ is the proportion of the number of different groups ). However, if the model learns how to classify sensitive groups, it leads to:

During inference, a group fairness model achieves parity in predictive performance across different sensitive groups. As a result, it predicts all cases for $k$ would have :

\begin{equation}
\scalebox{0.93}[1]{$TP_A = \alpha TP_{A'}, FP_A = \alpha FP_{A'}, TN_A = \alpha TN_{A'}, FN_A = \alpha FN_{A'}$}
\end{equation}

Where $\alpha$ is the proportion of the total number difference between different sensitive groups (e.g., if the total number of group $A$ is two times more than group $A'$, then $\alpha$ is 2), the $TP_A$, $FP_A$, $TN_A$, and $FN_A$ represent true positive cases, false positive cases, true negative cases, and false negative cases in the prediction for group $A$. In contrast, the subscripts $A'$ represent the prediction case for group $A'$. However, if the model learns how to classify sensitive groups, it leads to:

% \begin{enumerate}
%     \item \(FP_A > FP_{A'}, FP_A = FP_{A'} + X\) ($X$ is the increase in FP cases due to bias), \(TN_A = TN_{A'} - X\).
%     \item \(FN_{A'} > FN_A, FN_{A'} = FN_A + Y\) ($Y$ is the increase in FN cases due to bias), \(TP_{A'} = TP_A - Y\).
% \end{enumerate}

\begin{enumerate}
    \item \(FP_A > \alpha FP_{A'}, FP_A = \alpha FP_{A'} + X\) ($X$ is the increase in FP cases due to bias, $X > 0$), \(TN_A = \alpha TN_{A'} - X\).
    \item \(\alpha FN_{A'} > FN_A, \alpha FN_{A'} = FN_A + Y\) ($Y$ is the increase in FN cases due to bias, $Y > 0$), \(\alpha TP_{A'} = TP_A - Y\).
\end{enumerate}

Based on Eq.(\ref{eq.EOpp0_equation}), the values of Eopp0 for class $k$ can be calculated as follow :

% \begin{equation}
% \label{eq.eopp0_ref}
%     Eopp0 = \scalebox{1.13}[1]{$|TNR_{k}^{A} - TNR_{k}^{A'}| = \left|\frac{TN_{A}}{TN_{A}+FP_{A}} - \frac{TN_{A'}}{TN_{A'}+FP_{A'}}\right|$}
% \end{equation}

\begin{equation}
\label{eq.eopp0_ref}
\begin{split}
    Eopp0 &= \scalebox{1.05}[1]{$|TNR_{k}^{A} - TNR_{k}^{A'}| = \left|\frac{TN_{A}}{TN_{A}+FP_{A}} - \frac{TN_{A'}}{TN_{A'}+FP_{A'}}\right|$} \\& = \left|\frac{\alpha TN_{A'} - X}{\alpha TN_{A'}+\alpha FP_{A'}} - \frac{TN_{A'}}{TN_{A'}+FP_{A'}}\right| \\& = \left|\frac{TN_{A'} - \frac{X}{\alpha}}{TN_{A'}+FP_{A'}} - \frac{TN_{A'}}{TN_{A'}+FP_{A'}}\right| \\& = \left| - \frac{\frac{X}{\alpha}}{TN_{A'}+FP_{A'}} \right| \\& = \frac{X}{\alpha(TN_{A'}+FP_{A'})} = \frac{X}{TN_{A}+FP_{A}}
\end{split}
\end{equation}

% Our designed framework, which contains a regularization term, can make the model classify the class only based on the features related to the target attribute and prevent the model from using sensitive attribute information for classifying the class. Then, $X$ will decrease because $FP$ cases in prediction decrease, leading $FP_{A}$$\downarrow$, $TN_{A}$$\uparrow$, which make the values of $Eopp0$ decrease.

Our designed framework, which contains a regularization term, can make the model classify the class only based on the features related to the target attribute and prevent the model from using sensitive attribute information for classifying the class. Then, $X$ will decrease because $FP$ cases from shortcut prediction decrease, which decreases the values of $Eopp0$.

Next, we can calculate the Eopp1 based on Eq.(\ref{eq.EOpp1_equation}) as follow:

% \begin{equation}
%     Eopp1 = \scalebox{1.13}[1]{$|TPR_{k}^{A} - TPR_{k}^{A'}| = \left|\frac{TP_{A}}{TP_{A}+FN_{A}} - \frac{TP_{A'}}{TP_{A'}+FN_{A'}}\right|$}
% \end{equation}

\begin{equation}
\label{eq.eopp1_ref}
\begin{split}
    Eopp1 &= \scalebox{1.05}[1]{$|TPR_{k}^{A} - TPR_{k}^{A'}| = \left|\frac{TP_{A}}{TP_{A}+FN_{A}} - \frac{TP_{A'}}{TP_{A'}+FN_{A'}}\right|$} \\& = \left|\frac{\alpha TP_{A'} + Y}{\alpha TP_{A'}+\alpha FN_{A'}} - \frac{TP_{A'}}{TP_{A'}+FN_{A'}}\right| \\& = \left|\frac{TP_{A'} + \frac{Y}{\alpha}}{TP_{A'} + FN_{A'}} - \frac{TP_{A'}}{TP_{A'}+FN_{A'}}\right| \\& = \left| \frac{\frac{Y}{\alpha}}{TP_{A'}+FN_{A'}} \right| \\& = \frac{Y}{\alpha(TP_{A'}+FN_{A'})} = \frac{Y}{TP_{A}+FN_{A}}
\end{split}
\end{equation}

% Similar to the condition described in Eq.(\ref{eq.eopp0_ref}), if we apply the regularization term, which can make the model only classify the class based on the features related to the target attribute, then $Y$ will decrease, because $FN$ cases in prediction decrease, and then $FN_{A'}$$\downarrow$, $TP_A'$$\uparrow$, leading to the reduction of the $Eopp1$ value. 

Similar to the condition described in Eq.(\ref{eq.eopp0_ref}), if we apply the regularization term, which can make the model only classify the class based on the features related to the target attribute, then $Y$ will decrease, leading to the reduction of the $Eopp1$ value. 

As for the values of Eodd, we can calculate as follows based on Eq.(\ref{eq.EO_equation}): 

% \begin{equation}
% \begin{gathered}
%     Eodd = |(TPR_{k}^{A} - TPR_{k}^{A'}) + (FPR_{k}^{A} - FPR_{k}^{A'})| \\
%     \scalebox{1.13}[1]{$= \left|(\frac{TP_{A}}{TP_{A}+FN_{A}} - \frac{TP_{A'}}{TP_{A'}+FN_{A'}}) + (\frac{FP_{A}}{FP_{A}+TN_{A}} - \frac{FP_{A'}}{FP_{A'}+TN_{A'}})\right|$}
% \end{gathered}
% \end{equation}

\begin{equation}
% \begin{gathered}
\begin{split}
    & Eodd = |TPR_{k}^{A} - TPR_{k}^{A'} + FPR_{k}^{A} - FPR_{k}^{A'}| \\&
    \scalebox{1.00}[1]{$= \left|\frac{TP_{A}}{TP_{A}+FN_{A}} - \frac{TP_{A'}}{TP_{A'}+FN_{A'}} + \frac{FP_{A}}{FP_{A}+TN_{A}} - \frac{FP_{A'}}{FP_{A'}+TN_{A'}}\right|$} \\&
    \scalebox{1.00}[1]{$= \left|\frac{Y}{\alpha(TP_{A'}+FN_{A'})} + \frac{\alpha FP_{A'} + X}{\alpha FP_{A'}+ \alpha TN_{A'}} - \frac{FP_{A'}}{FP_{A'}+TN_{A'}}\right|$} \\&
    \scalebox{1.00}[1]{$= \left|\frac{Y}{\alpha(TP_{A'}+FN_{A'})} + \frac{FP_{A'} + \frac{X}{\alpha}}{FP_{A'}+ TN_{A'}} - \frac{FP_{A'}}{FP_{A'}+TN_{A'}}\right|$} \\&
    \scalebox{1.00}[1]{$= \left|\frac{Y}{\alpha(TP_{A'}+FN_{A'})} + \frac{X}{\alpha(FP_{A'}+TN_{A'})} \right|$} 
    \\& \scalebox{1.00}[1]{$= \frac{Y}{TP_{A}+FN_{A}} + \frac{X}{FP_{A}+TN_{A}}$} 
\end{split}
% \end{gathered}
\end{equation}

% Finally, similar to the above circumstance, the regularization term in our framework can make both $X$ and $Y$ decrease because both $FP$ and $FN$ cases decrease, which leads to $FP_{A}$$\downarrow$, $TN_{A}$$\uparrow$ and $FN_{A'}$$\downarrow$, $TP_A'$$\uparrow$, leading to $Eodd$ value decrease.

Finally, similar to the above circumstance, the regularization term in our framework can make both $X$ and $Y$ decrease because both $FP$ and $FN$ cases from shortcut prediction decrease, which leads to $Eodd$ value decrease.

Based on the decrease in all fairness metrics, it shows that avoiding the model to learn the biased sensitive attribute information can mitigate the shortcuts in model learning and then enhance the fairness of classification.

\begin{figure*}[!h]
\setlength{\belowcaptionskip}{-15pt} 
\begin{center}

\includegraphics[width=1.0\linewidth]{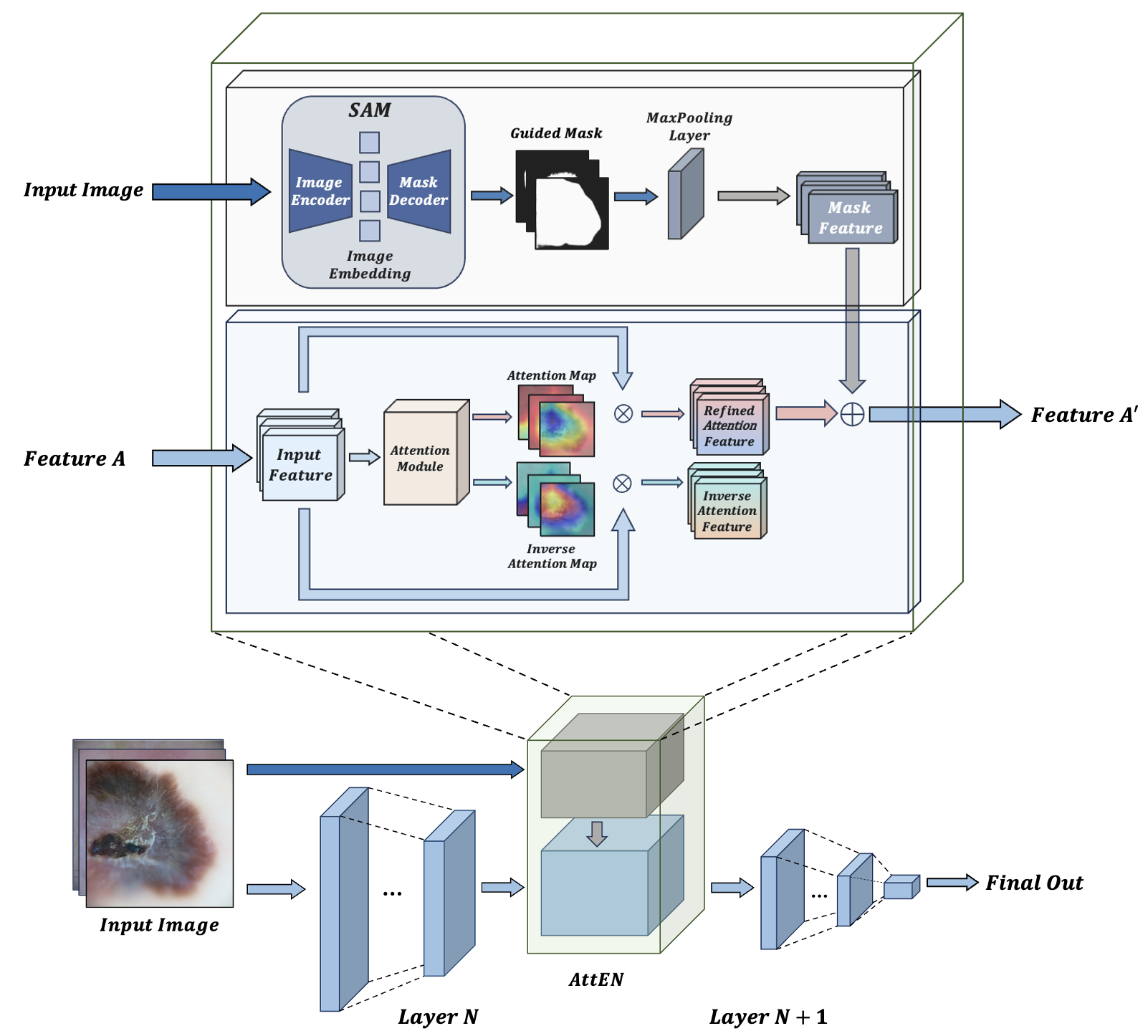}
\caption{Illustration of the proposed ``AttEN'' training framework. The upper arrow entering the AttEN module represents the original input images, while the lower arrow entering AttEN represents the features from previous internal layers, denoted as \textit{Feature $A$}. The arrow from the AttEN module, marked as \textit{Feature $A^{'}$}, will proceed to the next layer in the model. This module can be appended at any layer in the neural network; it can also append multiple modules in the neural network.
}
\label{Fig.Method}
\end{center}
\end{figure*}

\subsection{Training Framework}

In this paper, we aim to introduce a method to ensure fair predictions even when sensitive attribute information cannot be accessed. We propose a module called ``AttEN'', which can be embedded in the model's inner layer. The main idea for this module is to divide the dermatological disease image features learned by the model into two parts: disease and skin. When distinguishing disease categories, we aim to ensure that the model focuses on using features related to the disease and does not differentiate features related to the skin under the same condition. It is because demographic information is related to skin information (e.g., skin tone), and this approach can prevent bias in predictions based on skin features, especially in cases where there is a demographic imbalance within the same disease category.

As shown in Fig. \ref{Fig.Method}, this module includes an attention module that helps capture features related to the disease and skin. We regularize the features using the SNNL based on whether they belong to the disease or skin attributes. This allows the model to learn how to classify diseases based on disease-relevant features rather than skin-relevant features. To further improve the quality of the attention map generated from the attention module, we add additional information from the segmentation mask generated by SAM; we refer to it as the ``Guided Mask" in our framework. Our motivation is drawn from \cite{wang2021mask}. However, because the confidence in the masks generated by SAM is not always high, instead of directly using them as ground truth for the total loss calculation, as done in \cite{wang2021mask}, we incorporate these features into the final features as additional information to enhance feature quality. Specifically, we resize these masks to tensors that match the current input feature size through max pooling. These tensors guide the attention map and add to the features that focus on the diseased part generated from the attention module; then, the refined features are passed to the next layer. We will demonstrate the importance of this guidance in Section \ref{sec:results}.

% \begin{figure*}[!h]
% \setlength{\belowcaptionskip}{-15pt} 
% \begin{center}

% \includegraphics[width=1.0\linewidth]{Fig/Atten_Method.png}
% \caption{Illustration of the proposed ``AttEN'' training framework. The upper arrow entering the AttEN module represents the original input images, while the lower arrow entering AttEN represents the features from previous internal layers, denoted as \textit{Feature $A$}. The arrow from the AttEN module, marked as \textit{Feature $A^{'}$}, will proceed to the next layer in the model. This module can be appended at any layer in the neural network; it can also append multiple modules in the neural network.
% }
% \label{Fig.Method}
% \end{center}
% \end{figure*}

\subsection{Details for the Proposed AttEN Module}
This section will discuss the approach to implementing our proposed ``AttEN'' module. Include the implementation for the attention module and the formulation for the loss function, which can regularize the feature model learned in the training time.

\subsubsection{Attention Module}
To obtain features that focus on the diseased part or the skin part, we achieve our goal through an attention module. We adopt CBAM \cite{woo2018cbam}, a simple yet effective attention module that can be integrated into CNNs, as our attention module block. This module consists of two sequential sub-modules, the Channel Attention Module and Spatial Attention Module. The module can generate attention maps for input features, capturing more disease-related features as the model learns to classify diseases. Conversely, if we invert the generated attention maps, they will focus more on regions outside the diseased area, helping us create features more related to the skin part. The attention maps generated by CBAM are multiplied by the input feature map, producing the ``Refined Attention Feature", as shown in Fig. \ref{Fig.Method}. The inverse attention map, which focuses on the skin part, is multiplied by the original input feature to generate the ``Inverse Attention Feature". We then calculate SNNL for these features according to our training framework, followed by subsequent regularization loss calculations. 

\subsubsection{Loss Function}
The soft nearest neighbor loss (SNNL), as introduced in \cite{frosst2019analyzing}, is used to measure the degree of entanglement between features for different classes in the embedded space.

\textbf{Definition.} The soft nearest neighbor computes the features' entanglement degree, where $x$ are the features and $y$ is the label that can be either the target or sensitive attribute with batch size $b$.
\begin{equation}
\label{eq.snnl}
  L_{s n}(x, y, T)=-\frac{1}{b} \sum_{i \in 1 . . b} \log \left(\frac{\sum_{\substack{j \in 1 . . b \\ j \neq i \\ y_i=y_j}} e^{-\frac{\left\|x_i-x_j\right\|^2}{T}}}{\sum_{\substack{\in 1 . . b \\ k \neq i}} e^{-\frac{\left\|x_i-x_k\right\|^2}{T}}}\right)
\end{equation}

In our problem, since sensitive attributes are entirely inaccessible, we differentiate features based on their focus on skin or disease parts rather than evaluating feature entanglement based on sensitive attributes or target attributes. According to our training framework, for features that focus on the disease part, we need to increase the entanglement of features from the same disease category and decrease the entanglement of features from different disease categories. Therefore, when calculating SNNL using Eq.(\ref{eq.snnl_d}), where $x$ are the features focused on the diseased part and $y$ are the corresponding target labels to the features with temperature $T$. We compute SNNL for these disease-related features once for the batch and use this term as a regularization term to minimize during training, thus achieving this goal.

\begin{equation}
\label{eq.snnl_d}
  l_{disease} = L_{s n}(x, y, T)
\end{equation}

As for features focusing on the skin part, we aim to ensure that the differences in skin among the same disease category do not affect the prediction results. Therefore, based on Eq.(\ref{eq.snnl_s}), where $x^{'}$ are the features focused on the skin part and $y_{c}$ are the corresponding target labels with temperature $T$. We calculate SNNL, where $c$ represents the disease class and $C$ represents the total number of the classes, and by introducing this term as a regularization term during training and maximizing it, we can enhance the entanglement between skin features within the same disease category. In other words, the model will be unable to distinguish differences in the skin part under the same disease condition and will make judgments based on the disease part. By doing this, we can ensure fairness in diagnosis.

\begin{equation}
\label{eq.snnl_s}
  l_{skin} = \sum_{c=1}^{C} L_{s n}(x_{c}', y_{c}, T)
\end{equation}

It's worth noting that, in the definition of Eq.(\ref{eq.snnl}), if we select elements of the same class from the same batch when calculating SNNL, the result of Eq.(\ref{eq.snnl_s}) will be 0. To address this issue, during the implementation of SNNL calculation, for the numerator part, we do not exclude the case when $j \neq i$. Consequently, this ensures that the numerator is always greater than the denominator. This implementation technique ensures that even when obtaining features of the same class, the loss continues to be maximized, reducing the distance between features of the same class.

Based on the Eq.(\ref{eq.snnl_d}) and Eq.(\ref{eq.snnl_s}), the final loss used for the model optimization can be defined as follows, where $N$ represents the total number of the ``AttEN'' module append into the neural network, the $l_{disease}^{i}$ and $l_{skin}^{i}$ represent the loss function described in Eq.(\ref{eq.snnl_d}) and Eq.(\ref{eq.snnl_s}) which applied to each ``AttEN'' module, and the $l_{ce}$ is the cross entropy loss:

\begin{equation}
  loss = l_{ce} + \sum_{i=1}^{N} \alpha_{i} \left( l_{\!disease}^{i} - l_{\!skin}^{i} \right)
\end{equation}

With this training framework and loss function, the model can learn how to classify the disease based on the diseased part of the image, which can preserve predicted accuracy and improve fairness at the same time.

\section{Experiments}
\label{sec:exp}
\subsection{Dataset} 
In this study, we evaluate our method on two dermatological disease datasets, including ISIC 2019 challenge \cite{combalia2019bcn20000,tschandl2018ham10000} and the Fitzpatrick-17k dataset \cite{groh2021evaluating}. ISIC 2019 challenge contains 25,331 images in 8 diagnostic categories for target labels, and we take gender as our sensitive attribute. The Fitzpatrick-17k dataset contains 16,577 images in 114 skin conditions of target labels, following \cite{wu2022fairprune, chiu2023toward}, we categorize the skin tones into two groups, where types 1 to 3 are light skin, and types 4 to 6 are dark skin, and define it as the sensitive attribute. Our proposed method will not access these binary-sensitive attributes during training and validation time and will only use them to evaluate fairness performance during testing. Subsequently, we apply data augmentation techniques, including random flipping, rotation, scaling, and autoaugment as proposed by \cite{cubuk2018autoaugment}. Following this, we employ the stratified split approach, same as \cite{wu2022fairprune}, which split concerning both sensitive attribute and target attribute for ISIC 2019 dataset and only for target attribute for Fitzpatrick-17k dataset due to the data is not enough to do a stratified split for both attributes. The dataset is split into train, validation, and test with a ratio of 6:2:2. We repeat this process three times to ensure consistency and calculate both the average and standard deviation of the results, thereby enhancing the reliability of our method.

% 這邊的stratified split要怎麼寫

\begin{table*}[h]
\centering
\caption{\textmd{Results of accuracy and fairness of different methods on \textit{ISIC 2019} dataset, using gender as the sensitive attribute. The female is the privileged group with higher accuracy by vanilla training (Following most fairness works \cite{bellamy2018ai, du2021fairness}, we name the groups with advantages or desired model outcomes, typically with higher accuracy, as privileged groups.). For clarity, $Eopp0$ is abbreviated to $(\times 10^{-2})$, and $Eopp1$, $Eodd$, and $FATE$ are abbreviated to $(\times 10^{-1})$. \textbf{\underline{Best}} fairness results in method w/o sensitive attribute and \textbf{Best} fairness results in method w/ sensitive attribute are highlighted.}}
% \textbf{\underline{Best}} in method w/o sensitive attribute and \textbf{Best} in method w/ sensitive attribute are highlighted.

% All the model backbones employed in this study are based on \textit{ResNet18}.

\setlength{\tabcolsep}{10pt}
\label{Table:compare_isic}
\scriptsize
\begin{tabular}{c c | c c c | c c c}\toprule 

\multicolumn{1}{c}{\text{}} & \multicolumn{1}{c}{\text{}} &\multicolumn{3}{c}{\text{Accuracy}} & \multicolumn{3}{c}{\text{Fairness}}  \\ 
\cmidrule(lr){3-5}
\cmidrule(ll){6-8}
\multicolumn{1}{c}{\text{Method}} & \text{Gender} &\text{Precision} & \text{Recall} & \text{F1-score} & \text{Eopp0 $\downarrow$ / FATE $\uparrow$} & \text{Eopp1 $\downarrow$ / FATE $\uparrow$} & \text{Eodd $\downarrow$ / FATE $\uparrow$} \\ 
% \hline
\hline
 {\multirow{2}{*}{ResNet18}} & \text{Female} & 0.786±0.018 & 0.745±0.021 & 0.760±0.015 & \multirow{2}{*}{0.58±0.16 / 0.00±0.00} & \multirow{2}{*}{0.54±0.09 / 0.00±0.00} & \multirow{2}{*}{0.28±0.06 / 0.00±0.00} \\ 
 & \text{Male} & 0.753±0.030 & 0.757±0.030 & 0.753±0.031 &  &  & \\ 

\hline
\hline
\multicolumn{8}{c}{Methods w/ Sensitive Attribute Information} \\
\hline
\hline
\multicolumn{1}{ c }{\multirow{2}{*}{MFD}} & \text{Female} & 0.786±0.015 & 0.724±0.023 & 0.749±0.018 & \multirow{2}{*}{\textbf{0.51±0.03} / \textbf{0.91±1.94}} & \multirow{2}{*}{0.55±0.05 / -0.19±1.43} & \multirow{2}{*}{0.27±0.04 / 0.42±1.59} \\ 
 & \text{Male} & 0.802±0.031 & 0.758±0.028 & 0.775±0.027 &  &  & \\ 
\hline

{\multirow{2}{*}{FairPrune}} & \text{Female} & 0.743±0.030 & 0.718±0.011 & 0.724±0.009 & \multirow{2}{*}{0.54±0.20 / 0.23±1.84} & \multirow{2}{*}{0.44±0.17 / 1.61±2.52} & \multirow{2}{*}{0.23±0.08 / 1.66±1.96} \\ 
 & \text{Male} & 0.718±0.031 & 0.733±0.014 & 0.718±0.021 &  &  & \\ 
\hline

\multicolumn{1}{ c }{\multirow{2}{*}{FairAdaBN}} & \text{Female} & 0.784±0.015 & 0.729±0.012 & 0.752±0.012 & \multirow{2}{*}{0.51±0.07 / 0.79±1.53} & \multirow{2}{*}{\textbf{0.40±0.08 / 2.64±0.60}} & \multirow{2}{*}{\textbf{0.20±0.04 / 2.75±0.66}} \\ 
 & \text{Male} & 0.768±0.002 & 0.730±0.004 & 0.745±0.002 &  &  & \\ 
\hline
\hline
\multicolumn{8}{c}{Methods w/o Sensitive Attribute Information} \\
\hline
\hline

\multicolumn{1}{ c }{\multirow{2}{*}{SSLwD}} & \text{Female} & 0.791±0.046 & 0.669±0.041 & 0.716±0.042 & \multirow{2}{*}{0.68±0.10 / -3.24±5.26} & \multirow{2}{*}{0.41±0.07 / 1.91±0.80} & \multirow{2}{*}{0.22±0.04 / 1.63±1.62} \\ 
 & \text{Male} & 0.790±0.046 & 0.684±0.048 & 0.726±0.047 &  &  & \\ 
\hline

 \multicolumn{1}{ c }{\multirow{2}{*}{FDKD}} & \text{Female} & 0.759±0.033 & 0.713±0.032 & 0.731±0.031 & \multirow{2}{*}{0.57±0.05 / -0.89±3.70} & \multirow{2}{*}{0.46±0.21 / 1.14±3.54} & \multirow{2}{*}{0.25±0.10 / 0.69±3.33} \\ 
 & \text{Male} & 0.744±0.020 & 0.729±0.047 & 0.733±0.033 &  &  & \\ 
\hline

 \multicolumn{1}{ c }{\multirow{2}{*}{ME-ResNet18}} & \text{Female} & 0.752±0.004 & 0.770±0.040 & 0.758±0.021 & \multirow{2}{*}{0.73±0.12 / -3.66±5.64} & \multirow{2}{*}{0.43±0.11 / 2.05±0.89} & \multirow{2}{*}{0.22±0.06 / 2.10±0.81} \\ 
 & \text{Male} & 0.721±0.002 & 0.776±0.034 & 0.744±0.015 &  &  & \\ 

\hline
% \hline

\multicolumn{1}{ c }{\multirow{2}{*}{\textbf{AttEN (Ours)}}} & \text{Female} & 0.772±0.014 & 0.757±0.027 & 0.762±0.010 & \multirow{2}{*}{\textbf{\underline{0.54±0.09}} / \textbf{\underline{0.48±1.02}}} & \multirow{2}{*}{\textbf{\underline{0.27±0.10}} / \textbf{\underline{5.13±1.33}}} & \multirow{2}{*}{\textbf{\underline{0.15±0.06}} / \textbf{\underline{4.75±1.21}}} \\ 
 & \text{Male} & 0.741±0.026 & 0.759±0.020 & 0.747±0.015 &  &  & \\ 
 % \hline
\bottomrule
\end{tabular}
\end{table*}

\begin{table*}[h]
\centering
\caption{\textmd{Results of accuracy and fairness of different methods on \textit{Fitzpatrick-17k} dataset, using skin tone as the sensitive attribute. The dark skin is the privileged group with higher accuracy by vanilla training. For clarity, $Eopp0$ is abbreviated to $(\times 10^{-2})$, and $Eopp1$, $Eodd$, and $FATE$ are abbreviated to $(\times 10^{-1})$. \textbf{\underline{Best}} fairness results in method w/o sensitive attribute and \textbf{Best} fairness results in method w/ sensitive attribute are highlighted.}}

% All the model backbones employed in this study are based on \textit{ResNet18}.

\setlength{\tabcolsep}{10pt}
\label{Table:compare_fitz}
\scriptsize
\begin{tabular}{c c | c c c | c c c}\toprule 

\multicolumn{1}{c }{\text{}} & \multicolumn{1}{c }{\text{}} &\multicolumn{3}{c}{\text{Accuracy}} & \multicolumn{3}{c}{\text{Fairness}}  \\ 
\cmidrule(lr){3-5}
\cmidrule(ll){6-8}
\multicolumn{1}{ c }{\text{Method}} & \text{Skin Tone} &\text{Precision} & \text{Recall} & \text{F1-score} & \text{Eopp0 $\downarrow$ / FATE $\uparrow$} & \text{Eopp1 $\downarrow$ / FATE $\uparrow$} & \text{Eodd $\downarrow$ / FATE $\uparrow$} \\ 
\hline
\hline
 {\multirow{2}{*}{ResNet18}} & \text{Light} & 0.501±0.016 & 0.478±0.03 & 0.478±0.008 & \multirow{2}{*}{0.28±0.01 / 0.00±0.00} & \multirow{2}{*}{2.78±0.12 / 0.00±0.00} & \multirow{2}{*}{1.39±0.06 / 0.00±0.00} \\ 
 & \text{Dark} & 0.545±0.111 & 0.545±0.014 & 0.524±0.003 &  &  & \\ 

\hline
\hline
\multicolumn{8}{c}{Methods w/ Sensitive Attribute Information} \\
\hline
\hline

\multicolumn{1}{ c }{\multirow{2}{*}{MFD}} & \text{Light} & 0.522±0.021 & 0.481±0.003 & 0.487±0.008 & \multirow{2}{*}{0.28±0.01 / 0.23±0.75} & \multirow{2}{*}{2.68±0.12 / 0.59±0.06} & \multirow{2}{*}{1.34±0.06 / 0.58±0.07} \\ 
 & \text{Dark} & 0.575±0.005 & 0.551±0.003 & 0.539±0.003 &  &  & \\ 
\hline

{\multirow{2}{*}{FairPrune}} & \text{Light} & 0.504±0.049 & 0.478±0.012 & 0.477±0.025 & \multirow{2}{*}{\textbf{0.27±0.03} / \textbf{0.23±1.37}} & \multirow{2}{*}{\textbf{2.67±0.14} / 0.36±0.42} & \multirow{2}{*}{1.34±0.07 / 0.35±0.43} \\ 
 & \text{Dark} & 0.544±0.020 & 0.547±0.009 & 0.525±0.015 &  &  & \\ 
\hline

\multicolumn{1}{ c }{\multirow{2}{*}{FairAdaBN}} & \text{Light} & 0.529±0.004 & 0.485±0.007 & 0.493±0.006 & \multirow{2}{*}{0.28±0.02 / 0.11±0.99} & \multirow{2}{*}{2.68±0.07 / \textbf{0.81±0.57}} & \multirow{2}{*}{\textbf{1.30±0.04} / \textbf{0.79±0.57}} \\ 
 & \text{Dark} & 0.557±0.011 & 0.530±0.004 & 0.521±0.003 &  &  & \\ 
\hline
\hline
\multicolumn{8}{c}{Methods w/o Sensitive Attribute Information} \\
\hline
\hline

\multicolumn{1}{ c }{\multirow{2}{*}{SSLwD}} & \text{Light} & 0.495±0.031 & 0.454±0.023 & 0.462±0.025 & \multirow{2}{*}{0.29±0.02 / -0.94±1.13} & \multirow{2}{*}{2.60±0.08 / 0.31±0.12} & \multirow{2}{*}{1.30±0.04 / 0.30±0.12} \\ 
 & \text{Dark} & 0.537±0.026 & 0.519±0.012 & 0.506±0.019 &  &  & \\ 
\hline

 \multicolumn{1}{ c }{\multirow{2}{*}{FDKD}} & \text{Light} & 0.531±0.023 & 0.484±0.003 & 0.493±0.007 & \multirow{2}{*}{0.27±0.01 / 0.40±0.60} & \multirow{2}{*}{2.67±0.13 / 0.44±0.23} & \multirow{2}{*}{1.34±0.67 / 0.43±0.25} \\ 
 & \text{Dark} & 0.536±0.031 & 0.533±0.013 & 0.513±0.024 &  &  & \\ 
\hline

 \multicolumn{1}{ c }{\multirow{2}{*}{ME-ResNet18}} & \text{Light} & 0.520±0.023 & 0.479±0.006 & 0.481±0.012 & \multirow{2}{*}{0.28±0.01 / -0.01±0.54} & \multirow{2}{*}{2.53±0.04 / 0.99±0.27} & \multirow{2}{*}{1.27±0.02 / 0.99±0.28} \\ 
 & \text{Dark} & 0.558±0.014 & 0.552±0.015 & 0.533±0.007 &  &  & \\ 

\hline
% \hline

\multicolumn{1}{ c }{\multirow{2}{*}{\textbf{AttEN (Ours)}}} & \text{Light} & 0.526±0.010 & 0.483±0.009 & 0.492±0.010 & \multirow{2}{*}{\textbf{\underline{0.26±0.01}} / \textbf{\underline{0.75±0.45}}} & \multirow{2}{*}{\textbf{\underline{2.51±0.07}} / \textbf{\underline{1.09±0.17}}} & \multirow{2}{*}{\textbf{\underline{1.26±0.04}} / \textbf{\underline{1.09±0.15}}} \\ 
 & \text{Dark} & 0.555±0.020 & 0.539±0.009 & 0.524±0.012 &  &  & \\ 
 % \hline
\bottomrule
\end{tabular}
\end{table*}

\begin{table*}[ht]
\centering
\caption{\textmd{Results of accuracy and fairness of different methods on \textit{ISIC 2019} dataset, using gender as the sensitive attribute. The female is the privileged group with higher accuracy by vanilla training. For clarity, $Eopp0$ is abbreviated to $(\times 10^{-2})$, and $Eopp1$, $Eodd$, and $FATE$ are abbreviated to $(\times 10^{-1})$. \textbf{\underline{Best}} fairness results in method w/o sensitive attribute and \textbf{Best} fairness results in method w/ sensitive attribute are highlighted.}}

% All the model backbones employed in this study are based on \textit{VGG-11}. 
\setlength{\tabcolsep}{10pt}
\label{Table:compare_isic_vgg}
\scriptsize
\begin{tabular}{c c | c c c | c c c}\toprule 

\multicolumn{1}{c }{\text{}} & \multicolumn{1}{c }{\text{}} &\multicolumn{3}{c}{\text{Accuracy}} & \multicolumn{3}{c}{\text{Fairness}}  \\ 
\cmidrule(lr){3-5}
\cmidrule(ll){6-8}
\multicolumn{1}{ c }{\text{Method}} & \text{Gender} &\text{Precision} & \text{Recall} & \text{F1-score} & \text{Eopp0 $\downarrow$ / FATE $\uparrow$} & \text{Eopp1 $\downarrow$ / FATE $\uparrow$} & \text{Eodd $\downarrow$ / FATE $\uparrow$} \\ 
\hline
\hline
 {\multirow{2}{*}{VGG-11}} & \text{Female} & 0.750±0.029 & 0.711±0.024 & 0.726±0.012 & \multirow{2}{*}{0.71±0.18 / 0.00±0.00} & \multirow{2}{*}{0.61±0.12 / 0.00±0.00} & \multirow{2}{*}{0.43±0.12 / 0.00±0.00} \\ 
 & \text{Male} & 0.733±0.030 & 0.748±0.014 & 0.737±0.012 &  &  & \\ 

\hline
\hline
\multicolumn{8}{c}{Methods w/ Sensitive Attribute Information} \\
\hline
\hline
\multicolumn{1}{ c }{\multirow{2}{*}{MFD}} & \text{Female} & 0.751±0.006 & 0.732±0.034 & 0.735±0.023 & \multirow{2}{*}{\textbf{0.73±0.18} / \textbf{-0.99±4.40}} & \multirow{2}{*}{0.50±0.14 / 1.91±0.90} & \multirow{2}{*}{0.27±0.07 / 3.22±3.00} \\ 
 & \text{Male} & 0.738±0.012 & 0.741±0.021 & 0.734±0.018 &  &  & \\ 
\hline

{\multirow{2}{*}{FairPrune}} & \text{Female} & 0.750±0.029 & 0.709±0.030 & 0.722±0.015 & \multirow{2}{*}{0.76±0.16 / -0.99±1.81} & \multirow{2}{*}{0.49±0.25 / 2.19±2.86} & \multirow{2}{*}{0.32±0.05 / 1.93±2.27} \\ 
 & \text{Male} & 0.734±0.024 & 0.729±0.048 & 0.726±0.018 &  &  & \\ 
\hline

\multicolumn{1}{ c }{\multirow{2}{*}{FairAdaBN}} & \text{Female} & 0.701±0.008 & 0.733±0.006 & 0.714±0.003 & \multirow{2}{*}{0.91±0.06 / -3.44±2.27} & \multirow{2}{*}{\textbf{0.44±0.20 / 2.78±1.67}} & \multirow{2}{*}{\textbf{0.26±0.10 / 3.41±2.86}} \\ 
 & \text{Male} & 0.700±0.025 & 0.752±0.011 & 0.721±0.018 &  &  & \\ 
\hline
\hline
\multicolumn{8}{c}{Methods w/o Sensitive Attribute Information} \\
\hline
\hline

\multicolumn{1}{ c }{\multirow{2}{*}{SSLwD}} & \text{Female} & 0.806±0.021 & 0.671±0.022 & 0.719±0.016 & \multirow{2}{*}{0.68±0.01 / -0.03±2.21} & \multirow{2}{*}{0.45±0.09 / 2.50±0.40} & \multirow{2}{*}{0.25±0.05 / 3.63±2.27} \\ 
 & \text{Male} & 0.807±0.022 & 0.689±0.020 & 0.732±0.021 &  &  & \\ 
\hline

 \multicolumn{1}{ c }{\multirow{2}{*}{FDKD}} & \text{Female} &  0.743±0.014 & 0.728±0.007 & 0.732±0.007 & \multirow{2}{*}{0.68±0.30 / -0.53±5.89} & \multirow{2}{*}{0.44±0.13 / 2.83±0.85} & \multirow{2}{*}{0.25±0.06 / 3.91±2.28} \\ 
 & \text{Male} & 0.738±0.017 & 0.751±0.011 & 0.740±0.002 &  &  & \\ 
\hline

 \multicolumn{1}{ c }{\multirow{2}{*}{ME-VGG-11}} & \text{Female} & 0.717±0.046 & 0.722±0.027 & 0.711±0.042 & \multirow{2}{*}{\textbf{\underline{0.66±0.07}} / 0.33±1.66} & \multirow{2}{*}{0.45±0.13 / 2.46±1.79} & \multirow{2}{*}{0.24±0.07 / 3.75±3.56} \\ 
 & \text{Male} & 0.728±0.053 & 0.730±0.037 & 0.725±0.045 &  &  & \\ 

\hline
% \hline

\multicolumn{1}{ c }{\multirow{2}{*}{\textbf{AttEN (Ours)}}} & \text{Female} & 0.740±0.023 & 0.713±0.005 & 0.722±0.007 & \multirow{2}{*}{\textbf{\underline{0.66±0.07}} / \textbf{\underline{0.39±1.21}}} & \multirow{2}{*}{\textbf{\underline{0.38±0.11}} / \textbf{\underline{3.49±1.90}}} & \multirow{2}{*}{\textbf{\underline{0.20±0.06}} / \textbf{\underline{5.17±1.70}}} \\ 
 & \text{Male} & 0.730±0.030 & 0.729±0.015 & 0.727±0.023 &  &  & \\ 
 % \hline
\bottomrule
\end{tabular}
\end{table*}

\begin{table*}[h]
\centering
\caption{\textmd{Results of accuracy and fairness of different methods on \textit{Fitzpatrick-17k} dataset, using skin tone as the sensitive attribute. The dark skin is the privileged group with higher accuracy by vanilla training. For clarity, $Eopp0$ is abbreviated to $(\times 10^{-2})$, and $Eopp1$, $Eodd$, and $FATE$ are abbreviated to $(\times 10^{-1})$. \textbf{\underline{Best}} fairness results in method w/o sensitive attribute and \textbf{Best} fairness results in method w/ sensitive attribute are highlighted.}}

% All the model backbones employed in this study are based on \textit{VGG-11}.

\setlength{\tabcolsep}{10pt}
\label{Table:compare_fitz_vgg}
\scriptsize
\begin{tabular}{c c | c c c | c c c}\toprule 

\multicolumn{1}{c }{\text{}} & \multicolumn{1}{c }{\text{}} &\multicolumn{3}{c}{\text{Accuracy}} & \multicolumn{3}{c}{\text{Fairness}}  \\ 
\cmidrule(lr){3-5}
\cmidrule(ll){6-8}
\multicolumn{1}{ c }{\text{Method}} & \text{Skin Tone} &\text{Precision} & \text{Recall} & \text{F1-score} & \text{Eopp0 $\downarrow$ / FATE $\uparrow$} & \text{Eopp1 $\downarrow$ / FATE $\uparrow$} & \text{Eodd $\downarrow$ / FATE $\uparrow$} \\ 
\hline
\hline
 {\multirow{2}{*}{VGG-11}} & \text{Light} & 0.476±0.015 & 0.481±0.003 & 0.468±0.005 & \multirow{2}{*}{0.30±0.02 / 0.00±0.000} & \multirow{2}{*}{2.78±0.05 / 0.00±0.00} & \multirow{2}{*}{1.39±0.02 / 0.00±0.00} \\ 
 & \text{Dark} & 0.521±0.002 & 0.550±0.013 & 0.514±0.006 &  &  & \\ 

\hline
\hline
\multicolumn{8}{c}{Methods w/ Sensitive Attribute Information} \\
\hline
\hline
\multicolumn{1}{ c }{\multirow{2}{*}{MFD}} & \text{Light} & 0.478±0.005 & 0.494±0.007 & 0.474±0.007 & \multirow{2}{*}{0.30±0.02 / 0.16±1.35} & \multirow{2}{*}{2.65±0.09 / 0.67±0.38} & \multirow{2}{*}{1.33±0.05 / 0.66±0.38} \\ 
 & \text{Dark} & 0.544±0.028 & 0.556±0.025 & 0.527±0.024 &  &  & \\ 
\hline

{\multirow{2}{*}{FairPrune}} & \text{Light} & 0.485±0.023 & 0.474±0.005 & 0.466±0.006 & \multirow{2}{*}{0.29±0.01 / 0.34±0.52} & \multirow{2}{*}{\textbf{2.61±0.04} / 0.42±0.27} & \multirow{2}{*}{\textbf{1.31±0.02} / 0.41±0.28} \\ 
 & \text{Dark} & 0.525±0.003 & 0.532±0.031 & 0.496±0.021 &  &  & \\ 
\hline

\multicolumn{1}{ c }{\multirow{2}{*}{FairAdaBN}} & \text{Light} & 0.509±0.036 & 0.482±0.008 & 0.481±0.021 & \multirow{2}{*}{\textbf{0.28±0.01} / \textbf{0.95±0.30}} & \multirow{2}{*}{2.64±0.04 / \textbf{0.67±0.53}} & \multirow{2}{*}{1.32±0.02 / \textbf{0.66±0.53}} \\ 
 & \text{Dark} & 0.540±0.028 & 0.540±0.012 & 0.518±0.010 &  &  & \\ 
\hline
\hline
\multicolumn{8}{c}{Methods w/o Sensitive Attribute Information} \\
\hline
\hline

\multicolumn{1}{ c }{\multirow{2}{*}{SSLwD}} & \text{Light} & 0.465±0.041 & 0.447±0.042 & 0.441±0.033 & \multirow{2}{*}{0.33±0.04 / -1.89±0.86} & \multirow{2}{*}{2.71±0.19 / -0.37±0.29} & \multirow{2}{*}{1.35±0.10 / -0.37±0.28} \\ 
 & \text{Dark} & 0.531±0.003 & 0.517±0.038 & 0.499±0.014 &  &  & \\ 
\hline

 \multicolumn{1}{ c }{\multirow{2}{*}{FDKD}} & \text{Light} & 0.468±0.006 & 0.483±0.004 & 0.464±0.005 & \multirow{2}{*}{0.29±0.02 / 0.48±1.07} & \multirow{2}{*}{2.64±0.05 / 0.49±0.09} & \multirow{2}{*}{1.32±0.02 / 0.49±0.08} \\ 
 & \text{Dark} & 0.524±0.011 & 0.552±0.016 & 0.514±0.014 &  &  & \\ 
\hline

 \multicolumn{1}{ c }{\multirow{2}{*}{ME-VGG-11}} & \text{Light} & 0.462±0.006 & 0.461±0.014 & 0.447±0.010 & \multirow{2}{*}{0.30±0.01 / 0.00±0.88} & \multirow{2}{*}{\textbf{\underline{2.60±0.03}} / 0.46±0.22} & \multirow{2}{*}{1.30±0.02 / 0.46±0.23} \\ 
 & \text{Dark} & 0.535±0.018 & 0.550±0.016 & 0.515±0.013 &  &  & \\ 

\hline
% \hline

\multicolumn{1}{ c }{\multirow{2}{*}{\textbf{AttEN (Ours)}}} & \text{Light} & 0.478±0.010 & 0.478±0.004 & 0.467±0.005 & \multirow{2}{*}{\textbf{\underline{0.29±0.01}} / \textbf{\underline{0.55±0.64}}} & \multirow{2}{*}{2.60±0.10 / \textbf{\underline{0.80±0.17}}} & \multirow{2}{*}{\textbf{\underline{1.29±0.04}} / \textbf{\underline{0.84±0.14}}} \\ 
 & \text{Dark} & 0.538±0.025 & 0.565±0.011 & 0.528±0.017 &  &  & \\ 
 % \hline
\bottomrule
\end{tabular}
\end{table*}

\subsection{Evaluation Metrics}
To assess the fairness performance of our framework, we follow the previous work \cite{wu2022fairprune, chiu2023toward, xu2023fairadabn} and adopt the multi-class equalized opportunity metrics (Eopp0 and Eopp1) and the equalized odds metric (Eodd) proposed in \cite{hardt2016equality} for all experiments in Section \ref{sec:results}. Furthermore, to provide a more comprehensive assessment of our fairness performance, we employed the FATE metric introduced by \cite{xu2023fairadabn} to evaluate the balance between normalized fairness improvement and normalized accuracy drop. A higher FATE score indicates that the model achieves a more favorable trade-off between fairness and accuracy. The formula of FATE is shown below: 

\begin{equation}
\label{eq.FATE}
    FATE_{FC} = \frac{ACC_{m} - ACC_{b}}{ACC_{b}} - \lambda \frac{FC_{m} - FC_{b}}{FC_{b}}
\end{equation}

where $FC$ can take one of the values: Eopp0, Eopp1, or Eodd. $ACC$ denotes accuracy, and we measure accuracy using the F1-score for comparison. The subscripts $m$ and $b$ represent the bias mitigation and baseline models, respectively. The parameter $\lambda$ is utilized to adjust the importance of fairness in the final evaluation; here, we follow the setting of \cite{xu2023fairadabn} and set the $\lambda$ = 1.0 for simplification in the Section \ref{Compare_SOTA} and \ref{Compare_SOTA_vgg}.

\subsection{Implementation Details}
To ensure the generalizability of our method, we use ResNet18 and VGG-11 as the model backbones. The models are trained for 200 epochs using an SGD optimizer with a batch size 256 and a learning rate 1e-2. The backbone CNN consists of four internal ``AttEN'' modules inserted at the end of each residual block for ResNet18 and at the last four max-pooling layers for VGG-11, respectively. With this mechanism, the internal features extracted from each inference position are passed into the ``AttEN'' module to generate both the refined and inverse attention features required for SNNL computation.

% To ensure the generalizability of our method, we use ResNet18 and VGG-11 as the CNN backbones. The models are trained for 200 epochs using an SGD optimizer with a batch size of 256 and a learning rate 1e-2. The backbone CNN consists of four internal attention module blocks inserted at the end of each residual block for ResNet18 and at the last four max-pooling layers for VGG-11, respectively. In our implementation, we incorporate CBAM \cite{woo2018cbam}, a simple yet effective attention module that can be integrated into CNNs, as our attention module blocks. With this mechanism, the internal features extracted from each inference position are passed into the attention module to generate both the attention map and the inverse attention map required for soft nearest neighbor loss computation.

Regarding the generated guided mask, we employ the Medical SAM Adapter proposed by \cite{wu2023medical}. Since the ISIC 2019 and Fitzpatrick-17k datasets do not include ground truth segmentation masks, we fine-tuned the model on the ISIC 2018 dataset, which does provide segmentation ground truth masks. Following this fine-tuning, we apply this model to the ISIC 2019 and Fitzpatrick-17k datasets to generate the masks.

\begin{figure*}[h]
\begin{center}

% width=0.9\linewidth
\includegraphics[width=1.0\linewidth]{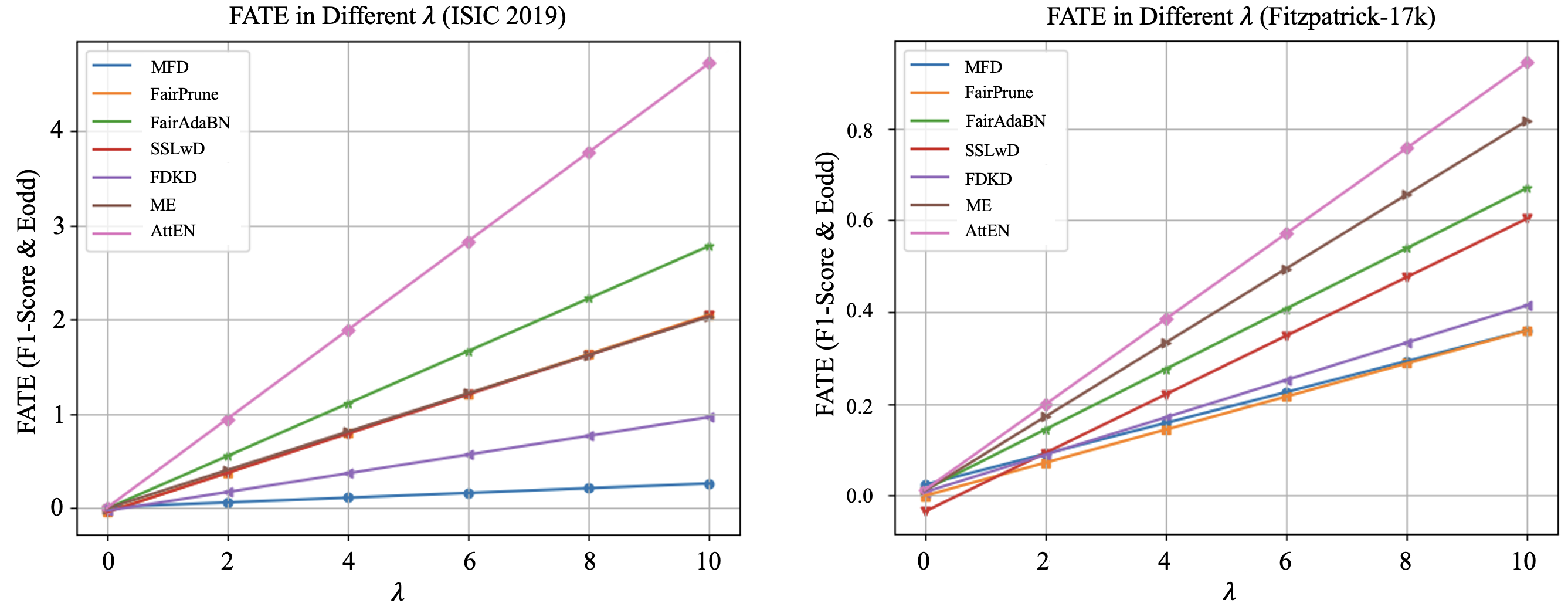}
\caption{The experimental results show the outcomes of the FATE metric under different $\lambda$ values. The horizontal axis represents the values of $\lambda$ ranging from 0 to 10. The vertical axis corresponds to the respective FATE values. In this comparison, we measure $FC$ by Eodd and $ACC$ by F1-Score, and the backbone model is ResNet18.
}
\label{Fig.lambda}
\end{center}
\end{figure*}

\section{Results}
\label{sec:results}

\subsection{Comparison with State-of-the-art}
\label{Compare_SOTA}
In this section, we compare our framework with several baselines, including CNN (ResNet18 \cite{he2016deep} and VGG-11 \cite{simonyan2014very}), MFD \cite{jung2021fair}, FairPrune \cite{wu2022fairprune}, FairAdaBN \cite{xu2023fairadabn}, SSLwD \cite{chai2022self}, FDKD \cite{chai2022fairness}, and ME-CNN (ME-ResNet18, ME-VGG-11) \cite{chiu2023toward}. MFD, FairPrune, and FairAdaBN require sensitive attribute information in the bias mitigation process, while SSLwD, FDKD, and ME-CNN can achieve fairness without knowledge of sensitive attributes. All baselines are reproduced by following the recommended hyperparameter settings of the original papers or resources. We report accuracy and fairness results for each dataset, including precision, recall, and F1-score metrics.

\textbf{ISIC 2019 Dataset.}
Table \ref{Table:compare_isic} displays the results of our method applied to ResNet18 running on the ISIC 2019 dataset. In this table, ``AttEN'' refers to our proposed method. Our approach has achieved the best fairness scores compared to methods that do not require sensitive attribute information. Compared with the baseline ResNet18, it shows an average decrease of 6.9\%, 50.0\%, and 46.4\% in Eopp0, Eopp1, and Eodd, respectively. It also shows the best FATE score across all methods, representing our proposed ``AttEN'' as having the best accuracy-fairness trade-off in the comparison. 

Moreover, when we compare our method to those that do necessitate sensitive attribute information, it consistently outperforms them in terms of Eopp1 and Eodd, showing 32.5\% and 25.0\% decrease compared with ``FairAdaBN'', which has the best score in these metrics. Our method also shows comparable Eopp0 score results compared to ``MFD''. However, it is worth emphasizing that our method does not rely on sensitive attribute information at any point, highlighting its capability to achieve the best overall improvement among all methods.

This achievement is primarily attributed to AttEN, which ensures that the model learns features that cannot distinguish disease types based on skin information. Instead, it enables the model to classify diseases solely based on the diseased part. This design ensures the same classification performance across different sensitive groups and improves the fairness score across different metrics.

\textbf{Fitzpatrick-17k Dataset.}
Table \ref{Table:compare_fitz} presents the results of our method applied to ResNet18 running on the Fitzpatrick-17k dataset. Our method outperforms other methods in all fairness metrics, including those using sensitive attributes. Compared to the best scores in Eopp0, Eopp1, and Eodd among methods using sensitive attributes, it shows an average decrease of 3.7\%, 5.9\%, and 3.1\%, respectively. Furthermore, our results demonstrate the best FATE score among all methods, highlighting the effectiveness of our method in achieving fairness even without the need for sensitive attributes.

\subsection{Comparison with State-of-the-art in Different Backbone}
\label{Compare_SOTA_vgg}

To further examine the generalizability of our model, we replaced our backbone model with VGG-11 and conducted the same experiments as Section \ref{Compare_SOTA}. The results are as follows. Table \ref{Table:compare_isic_vgg} and Table \ref{Table:compare_fitz_vgg} respectively represent the results of applying our method to VGG-11 and running it on the ISIC 2019 dataset and the Fitzpatrick-17k dataset. Our method shows the best results in all fairness metrics in the ISIC 2019 dataset. Compared to baseline VGG-11, it offers an average decrease of 7.0\%, 37.7\%, and 53.5\% in Eopp0, Eopp1, and Eodd, respectively. It also shows the best FATE score among all methods. As for the Fitzpatrick-17k dataset, our method showcases the best Eopp0 and Eodd scores and the best FATE score among all methods that do not require sensitive attributes. The experimental results highlight the extensibility of our method across different model backbones.

\subsection{The Impact for the Different Values of $\lambda$ on the FATE}

In the evaluation metric, FATE, the hyperparameters $\lambda$ is a weight factor that can affect the final results of the FATE value. To ensure that our method maintains favorable results under different weight settings, we calculated the FATE values for various $\lambda$ values in the range of 0 to 10 based on three different data splits and then averaged the results. As shown in Fig. \ref{Fig.lambda}, in the ISIC 2019 dataset, our AttEN method consistently achieves the highest score regardless of the $\lambda$ value, which means our method can get the best trade-off in the consideration for different weights in terms of fairness with accuracy. Similar results are observed in the Fitzpatrick-17k dataset, where our method consistently obtains the highest FATE values in most cases. These experiments can show our method's robustness trade-off performance in different considerations for fairness and accuracy.

\begin{table*}[h]
\centering
\caption{\textmd{Ablation study on whether to use a guided mask in the framework, using both dataset and \textit{ResNet18} as the model backbone in this study. For clarity, $Eopp0$ is abbreviated to $(\times 10^{-2})$, and $Eopp1$ and $Eodd$ are abbreviated to $(\times 10^{-1})$. \textbf{Best} fairness results are highlighted.}}
\setlength{\tabcolsep}{14pt}
\label{Table:ablaion_mask_merge}
\scriptsize
\begin{tabular}{c c | c c c | c c c}\toprule 
\multicolumn{1}{c}{\text{}} & \multicolumn{1}{c}{\text{}} &\multicolumn{3}{c}{\text{Accuracy}} & \multicolumn{3}{c}{\text{Fairness}}  \\ 
\cmidrule(lr){3-5}
\cmidrule(ll){6-8}
\multicolumn{1}{c}{\text{Method}} & \text{Sensitive attribute} &\text{Precision} & \text{Recall} & \text{F1-score} & \text{Eopp0 $\downarrow$} & \text{Eopp1 $\downarrow$} & \text{Eodd $\downarrow$} \\ 
% \hline
\hline
\hline
\multicolumn{8}{c}{ISIC 2019 dataset} \\
\hline
\hline
\multicolumn{1}{ c }{\multirow{2}{*}{\textbf{AttEN (w/ mask)}}} & \text{Female} & 0.772±0.014 & 0.757±0.027 & 0.762±0.010 & \multirow{2}{*}{\textbf{0.54±0.09}} & \multirow{2}{*}{\textbf{0.27±0.10}} & \multirow{2}{*}{\textbf{0.15±0.06}} \\ 
 & \text{Male} & 0.741±0.026 & 0.759±0.020 & 0.747±0.015 &  &  & \\ 
\hline
\multicolumn{1}{ c }{\multirow{2}{*}{AttEN (w/o mask)}} & \text{Female} & 0.764±0.006 & 0.731±0.013 & 0.745±0.007 & \multirow{2}{*}{0.55±0.11} & \multirow{2}{*}{0.38±0.17} & \multirow{2}{*}{0.20±0.07} \\ 
 & \text{Male} & 0.748±0.015 & 0.740±0.018 & 0.741±0.006 &  &  & \\ 
\hline
\hline
\multicolumn{8}{c}{Fitzpatrick-17k dataset} \\
\hline
\hline

\multicolumn{1}{ c }{\multirow{2}{*}{\textbf{AttEN (w/ mask)}}} & \text{Light} & 0.526±0.010 & 0.483±0.009 & 0.492±0.010 & \multirow{2}{*}{\textbf{0.26±0.01}} & \multirow{2}{*}{\textbf{2.51±0.07}} & \multirow{2}{*}{\textbf{1.26±0.04}} \\ 
 & \text{Dark} & 0.555±0.020 & 0.539±0.009 & 0.524±0.012 &  &  & \\  
\hline
\multicolumn{1}{ c }{\multirow{2}{*}{AttEN (w/o mask)}} & \text{Light} & 0.518±0.014 & 0.479±0.004 & 0.484±0.007 & \multirow{2}{*}{0.27±0.01} & \multirow{2}{*}{2.63±0.05} & \multirow{2}{*}{1.29±0.02} \\ 
 & \text{Dark} & 0.550±0.012 & 0.541±0.017 & 0.526±0.014 &  &  & \\ 
 % \hline
\bottomrule
\end{tabular}
\end{table*}

\begin{figure*}[!h]
\setlength{\belowcaptionskip}{-15pt} 
\begin{center}

% width=0.9\linewidth
\includegraphics[width=1.0\linewidth]{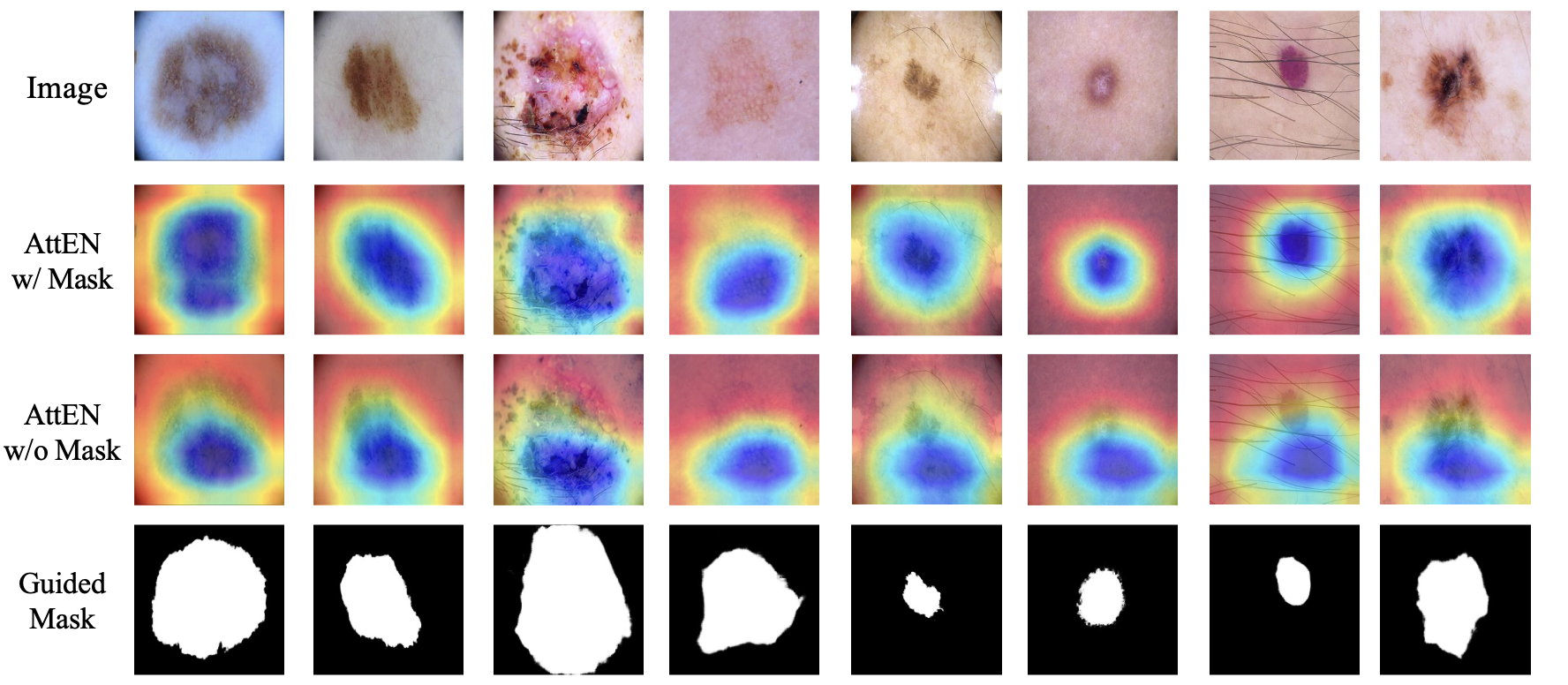}
\caption{The qualitative results show the Grad-CAM heatmap visualization for the ISIC 2019 dataset. The first row represents the original images, the second row denotes the AttEN method introducing the guided mask into the training process, the third row is the AttEN method without introducing the guided mask into the training process, and the last row is the row with the corresponding guided masks generated by SAM. Since the testing phase model does not take the guided masks as input, we put the guided masks here only for reference.}
\label{Fig.ablation_mask}
\end{center}
\end{figure*}

\subsection{Ablation Study}

\subsubsection{Ablation Study on Guided Mask}
In this section, we aim to investigate the impact of the guided mask and determine whether it can affect both fairness and accuracy results in the final predictions. We do the quantitative evaluation as Table \ref{Table:ablaion_mask_merge} shows; after incorporating the guided mask into the training framework, there is an improvement in all fairness metric scores for both datasets. Regarding the accuracy score, the results also demonstrate improvements for the Fitzpatrick-17k dataset in all metrics on average and the better F1-score in the ISIC 2019 dataset. 

We present one image for each class in the testing set of the ISIC 2019 dataset for the qualitative analysis, as shown in Fig. \ref{Fig.ablation_mask}. We leverage ResNet18 as the model backbone and base on the Grad-CAM \cite{selvaraju2017grad} to visualize the heatmap. It can be seen that with the guided mask, as shown in the second row, compared with the third row, which does not include the guided mask in the training framework, it has more accurate attention part on the image; that is, it puts more attention on disease part rather than skin part, avoid using skin information to predict the disease type. The quantitative and qualitative evaluation results demonstrate that the model can generate a more accurate attention map focusing on the diseased part by the guided mask information, improving accuracy and fairness performance. These results indicate the feasibility of our proposed method, which includes mask information in the training process.

\subsubsection{Ablation Study on Feature Entanglement}
To assess whether the technique that enhances feature entanglement across sensitive groups and reduces entanglement in different target groups is effective, we compared our proposed method with a method that only applies the attention module. As Table \ref{Table:ablaion_snnl_merge} illustrates, after incorporating feature entanglement regularization into the loss optimization, all fairness metric shows the improvement in both datasets, showing the 5.3\% and 3.7\% decrease in Eopp0, the 42.6\% and 2.0\% decrease in Eopp1 and the 40.0\% and 3.1\% decrease in Eodd in ISIC 2019 dataset and Fitzpatrick-17k dataset, respectively. This improvement occurs because our framework enables the model to distinguish features related to the disease. At the same time, it cannot differentiate between features related to different skin parts, which makes the model fairer because the sensitive attribute information of the image cannot affect the prediction result. These results support the effectiveness of our method in achieving fairness without relying on sensitive attributes in the dermatological disease task.

\begin{table*}[h]
\centering
\caption{\textmd{Ablation study on whether to use the soft nearest neighbor loss regularization in the framework, using both dataset and \textit{ResNet18} as the model backbone in this study. For clarity, $Eopp0$ is abbreviated to $(\times 10^{-2})$, $Eopp1$ and $Eodd$ are abbreviated to $(\times 10^{-1})$. \textbf{Best} fairness results are highlighted.}}
\setlength{\tabcolsep}{14pt}
\label{Table:ablaion_snnl_merge}
\scriptsize
\begin{tabular}{c c | c c c | c c c}\toprule 
\multicolumn{1}{c}{\text{}} & \multicolumn{1}{c}{\text{}} &\multicolumn{3}{c}{\text{Accuracy}} & \multicolumn{3}{c}{\text{Fairness}}  \\ 
\cmidrule(lr){3-5}
\cmidrule(ll){6-8}
\multicolumn{1}{c}{\text{Method}} & \text{Sensitive attribute} &\text{Precision} & \text{Recall} & \text{F1-score} & \text{Eopp0 $\downarrow$} & \text{Eopp1 $\downarrow$} & \text{Eodd $\downarrow$} \\ 
% \hline
\hline
\hline
\multicolumn{8}{c}{ISIC 2019 dataset} \\
\hline
\hline
\multicolumn{1}{ c }{\multirow{2}{*}{\textbf{AttEN (w/ snnl)}}} & \text{Female} & 0.772±0.014 & 0.757±0.027 & 0.762±0.010 & \multirow{2}{*}{\textbf{0.54±0.09}} & \multirow{2}{*}{\textbf{0.27±0.10}} & \multirow{2}{*}{\textbf{0.15±0.06}} \\ 
 & \text{Male} & 0.741±0.026 & 0.759±0.020 & 0.747±0.015 &  &  & \\ 
\hline
\multicolumn{1}{ c }{\multirow{2}{*}{AttEN (w/o snnl)}} & \text{Female} & 0.773±0.013 & 0.719±0.009 & 0.741±0.010 & \multirow{2}{*}{0.57±0.10} & \multirow{2}{*}{0.47±0.17} & \multirow{2}{*}{0.25±0.09} \\ 
 & \text{Male} & 0.787±0.010 & 0.745±0.009 & 0.758±0.016 &  &  & \\ 
\hline
\hline
\multicolumn{8}{c}{Fitzpatrick-17k dataset} \\
\hline
\hline

\multicolumn{1}{ c }{\multirow{2}{*}{\textbf{AttEN (w/ snnl)}}} & \text{Light} & 0.526±0.010 & 0.483±0.009 & 0.492±0.010 & \multirow{2}{*}{\textbf{0.26±0.01}} & \multirow{2}{*}{\textbf{2.51±0.07}} & \multirow{2}{*}{\textbf{1.26±0.04}} \\ 
 & \text{Dark} & 0.555±0.020 & 0.539±0.009 & 0.524±0.012 &  &  & \\ 
\hline
\multicolumn{1}{ c }{\multirow{2}{*}{AttEN (w/o snnl)}} & \text{Light} & 0.523±0.009 & 0.470±0.012 & 0.480±0.009 & \multirow{2}{*}{0.27±0.01} & \multirow{2}{*}{2.56±0.02} & \multirow{2}{*}{1.30±0.08} \\ 
 & \text{Dark} & 0.565±0.013 & 0.550±0.007 & 0.534±0.008 &  &  & \\ 
 % \hline
\bottomrule
\end{tabular}
\end{table*}

\section{Conclusion}
\label{sec:conclusion}
In this paper, we aim to mitigate prediction bias in dermatological disease diagnosis tasks without accessing sensitive attribute information. Our proposed ``AttEN'' module has an attention module that can capture the features related to the diseased and skin parts. These features can be regularized by SNNL, allowing the model to differentiate diseases based on features associated with the diseased part rather than the skin part, ensuring fair predictions among different sensitive groups. To further enhance the quality of the attention map learned by the module, we incorporate the guided mask generated by SAM into the training process and demonstrate the effectiveness of this design in the ablation study. We apply our ``AttEN'' module to different backbone models, showcasing its robust generalization capabilities. Our experimental results demonstrate that our framework achieves the best trade-off between accuracy and fairness compared to the state-of-the-art on two dermatological disease datasets.

% In conclusion, we have successfully mitigated prediction bias in dermatological disease diagnosis tasks without accessing sensitive attribute information. Our approach has been applied to different backbone models, showcasing its robust generalization capabilities. Our experimental results demonstrate that our framework achieves the best trade-off between accuracy and fairness compared to the state-of-the-art on two dermatological disease datasets.

% We address the issue of mitigating bias in dermatological disease diagnosis tasks without accessing sensitive attribute information. We apply our method to different backbone models to demonstrate its generalization. Our results demonstrate that our framework achieves the best trade-off between accuracy and fairness compared to the state-of-the-art on two dermatological disease datasets.

\bibliographystyle{IEEEtran}
\bibliography{conference_101719}

\end{document}